\def\year{2022}\relax
\documentclass[letterpaper]{article} 
\usepackage{aaai22}  
\usepackage{times}  
\usepackage{helvet}  
\usepackage{courier}  
\usepackage[hyphens]{url}  
\usepackage{color}
\usepackage{graphicx} 
\usepackage{natbib}  
\usepackage{caption} 
\DeclareCaptionStyle{ruled}{labelfont=normalfont,labelsep=colon,strut=off} 
\usepackage{algorithm}
\usepackage{algorithmic}
\usepackage{amsmath}
\usepackage[pagebackref,breaklinks,colorlinks,citecolor = blue]{hyperref}

\usepackage{graphicx}                  
\usepackage{float}
\usepackage{lscape}                                         
\usepackage{wrapfig}

\usepackage{booktabs}                   
\usepackage{multirow}
\usepackage{makecell}
\usepackage{paralist}
\usepackage{enumitem}

\usepackage{bm}                          
\usepackage{epsfig}                      
\usepackage{mathtools}
\usepackage{amssymb,amsmath}   

\usepackage{units}
\usepackage{color}
\usepackage{amsfonts}       
\usepackage{nicefrac}       
\usepackage{pifont}         
\usepackage[normalem]{ulem}

\usepackage{comment}

\usepackage{url}  
\usepackage{xspace}
\usepackage[table]{xcolor}
\usepackage{grfext}

\PrependGraphicsExtensions*{.jpg,.png,.PNG}

\usepackage[super]{nth}
\usepackage{array}
\usepackage{colortbl}

\usepackage{soul}
\usepackage[british,american]{babel} 
\usepackage{breakcites}
\usepackage{algorithm} 
\usepackage{algorithmic}

\newcolumntype{x}[1]{>{\centering\arraybackslash}p{#1pt}}
\newcolumntype{y}[1]{>{\raggedright\arraybackslash}p{#1pt}}
\newcolumntype{z}[1]{>{\raggedleft\arraybackslash}p{#1pt}}

\newlength\savewidth

\newcommand{\ignore}[1]{}   

\usepackage{xcolor}
\colorlet{dark-blue}{blue!50!black}
\colorlet{dark-cyan}{cyan!75!black}
\colorlet{dark-purple}{purple!50!black}
\colorlet{dark-red}{red!75!black}
\colorlet{dark-green}{green!75!black}
\colorlet{dark-orange}{orange!50!black}
\colorlet{dark-gray}{black!75}
\colorlet{light-gray}{black!30}
\definecolor{nice-red}{HTML}{E41A1C}
\definecolor{nice-orange}{HTML}{FF7F00}
\definecolor{nice-yellow}{HTML}{FFC020}
\definecolor{nice-green}{HTML}{39b54a}
\definecolor{nice-blue}{HTML}{0071bc}
\definecolor{nice-purple}{HTML}{984EA3}

\definecolor{darkGreen}{rgb}{0, 0.6, 0}
\definecolor{darkRed}{rgb}{0.9, 0, 0} 
\definecolor{cyan}{rgb}{0, 0.5, 0.6} 
\definecolor{darkViolet}{rgb}{0.58, 0, 0.83}

\definecolor{lightgreen}{rgb}{0.4, .9, 0.4}
\definecolor{lightred}{rgb}{1, 0.5, 0.51}
\definecolor{xgray}{rgb}{0.6, 0.6, 0.6}
\definecolor{Highlight}{HTML}{39b54a}
\definecolor{citecolor}{HTML}{0071bc}

\usepackage{xcolor}
\colorlet{dark-blue}{blue!50!black}
\usepackage{booktabs}
\usepackage{amssymb}
\usepackage{newfloat}
\usepackage{listings}
\floatstyle{ruled}
\newfloat{listing}{tb}{lst}{}
\floatname{listing}{Listing}
\title{Degrade is Upgrade: Learning Degradation for Low-light Image Enhancement}
\title{Degrade is Upgrade: Learning Degradation for Low-light Image Enhancement}
\author {
    Kui Jiang~\textsuperscript{\rm 1},
    Zhongyuan Wang~\textsuperscript{\rm 1}\thanks{Corresponding Author},
    Zheng Wang~\textsuperscript{\rm 1},
    Chen Chen~\textsuperscript{\rm 2}, 
    Peng Yi~\textsuperscript{\rm 1}, 
    Tao Lu~\textsuperscript{\rm 3}, 
    Chia-Wen Lin~\textsuperscript{\rm 4} 
}
\affiliations {
    \textsuperscript{\rm 1}~School of Computer Science, Wuhan University, Wuhan, China\\
    \textsuperscript{\rm 2}~University of Central Florida,
    \textsuperscript{\rm 3}~Wuhan Institute of Technology,
    \textsuperscript{\rm 4}~National Tsing Hua University \\
    \{kuijiang, wangzwhu, yipeng\}@whu.edu.cn, wzy\_hope@163.com, chen.chen@crcv.ucf.edu, lutxyl@gmail.com, cwlin@ee.nthu.edu.tw
}

\usepackage{bibentry}

\begin{document}

\maketitle

\begin{abstract}
Low-light image enhancement aims to improve an image's visibility while keeping its visual naturalness. Different from existing methods tending to accomplish the relighting task directly by ignoring the fidelity and naturalness recovery, we investigate the intrinsic degradation and relight the low-light image while refining the details and color in two steps. Inspired by the color image formulation (diffuse illumination color plus environment illumination color), we first estimate the degradation from low-light inputs to simulate the distortion of environment illumination color, and then refine the content to recover the loss of diffuse illumination color. To this end, we propose a novel Degradation-to-Refinement Generation Network (DRGN). Its distinctive features can be summarized as 1) A novel two-step generation network for degradation learning and content refinement. It is not only superior to one-step methods, but also capable of synthesizing sufficient paired samples to benefit the model training; 2) A multi-resolution fusion network to represent the target information (degradation or contents) in a multi-scale cooperative manner, which is more effective to address the complex unmixing problems. Extensive experiments on both the enhancement task and joint detection task have verified the effectiveness and efficiency of our proposed method, surpassing the SOTA by \textit{0.70dB on average and 3.18\% in mAP}, respectively. The code is available at \url{https://github.com/kuijiang0802/DRGN}.
\end{abstract}

\section{Introduction}
Low-light conditions cause a series of visibility degradation and even sometimes destroy the color or content of the image. Such signal distortion and detail loss often lead to the failure of many computer vision systems~\cite{Bae2019aaai,HuangLYWRH2018AAAI,yang2020mutualnet,yu2021towards,
xu2021rethinking,zhong2021grayscale,huang2021occluded,xu2021exploring}. Therefore, there is a pressing need to develop algorithms to produce normally exposed outputs.

Early low-light enhancement works mainly focus on contrast enhancement to recover the visibility of dark regions~\cite{land1977retinex}. Since the hand-crafted priors and assumptions are introduced for specific scenes or degradation conditions, these techniques~\cite{kaur2011survey} are less effective on scenarios when the predefined models do not hold. Recently, deep learning frameworks have emerged as a promising solution to low-light image enhancement~\cite{zhu2020eemefn, chen2018learning, zhang2021beyond, jiang2021enlightengan, zhu2020eemefn}. The existing methods can be roughly divided into two categories regarding the imaging models under low-light conditions.

\tabcolsep=0.5pt
\begin{figure}[t]
	\centering
	\renewcommand\arraystretch{0.3}
	\begin{tabular}{ccccc}
			\includegraphics[width=0.19\columnwidth]{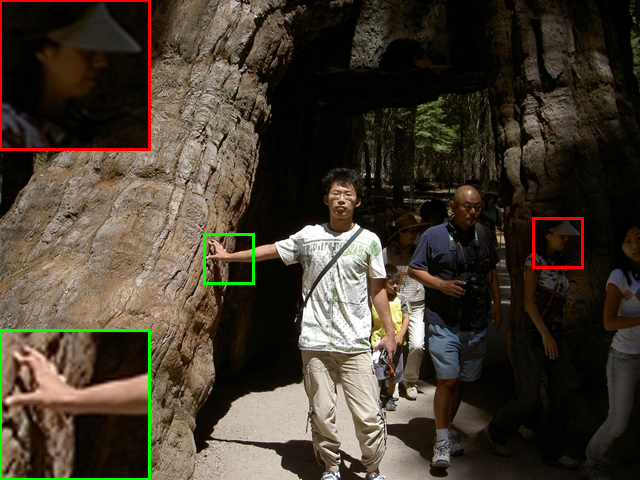} &
			\includegraphics[width=0.19\columnwidth]{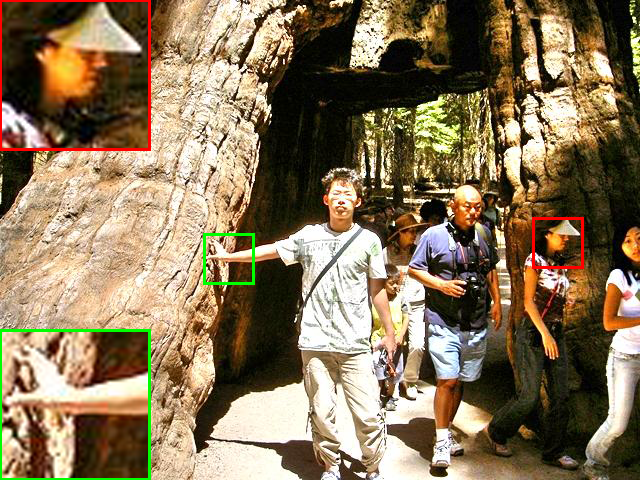} &
			\includegraphics[width=0.19\columnwidth]{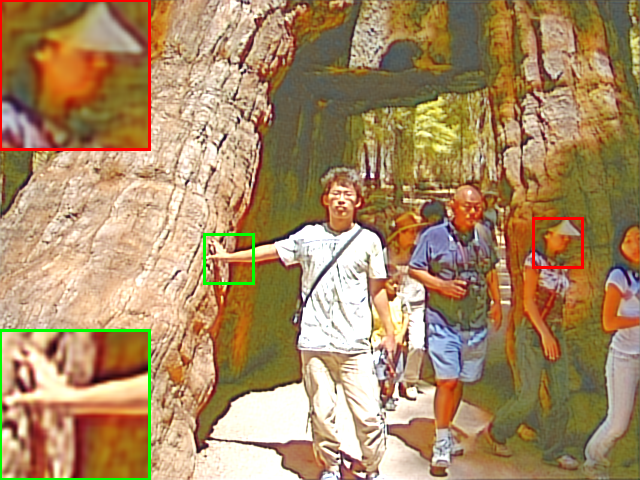} &
			\includegraphics[width=0.19\columnwidth]{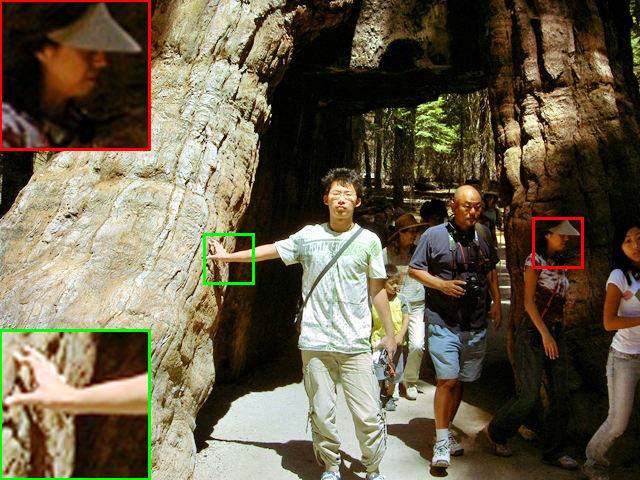} &
			\includegraphics[width=0.19\columnwidth]{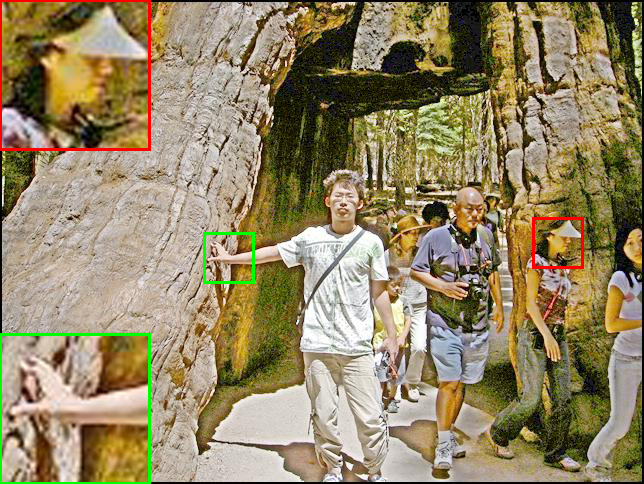} \\
			\scriptsize{Input} & \scriptsize{LIME} 
			& \scriptsize{RetinexNet}
			& \scriptsize{DeepUPE}
			& \scriptsize{Zero-DCE}
			\\
            \includegraphics[width=0.19\columnwidth]{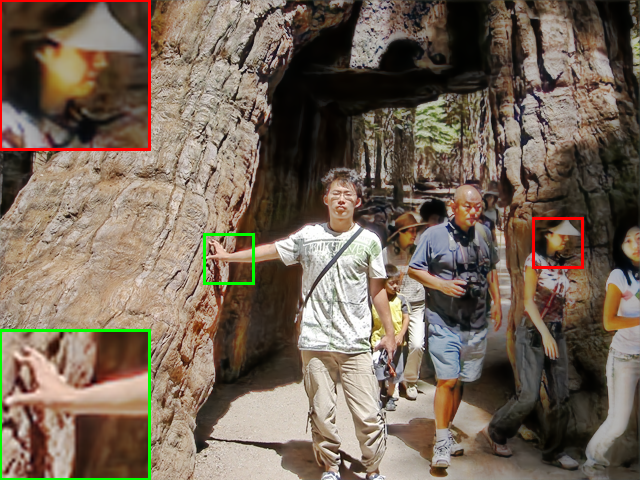} &
			\includegraphics[width=0.19\columnwidth]{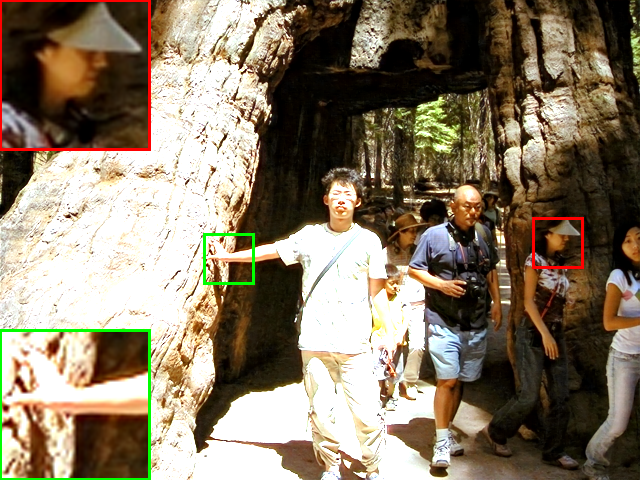} &
			\includegraphics[width=0.19\columnwidth]{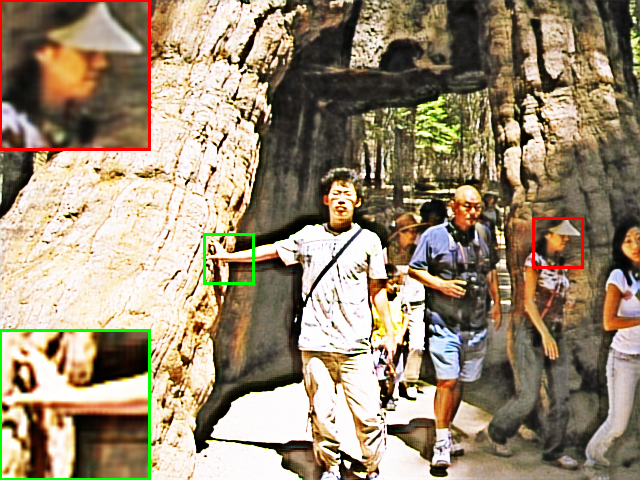} &
			\includegraphics[width=0.19\columnwidth]{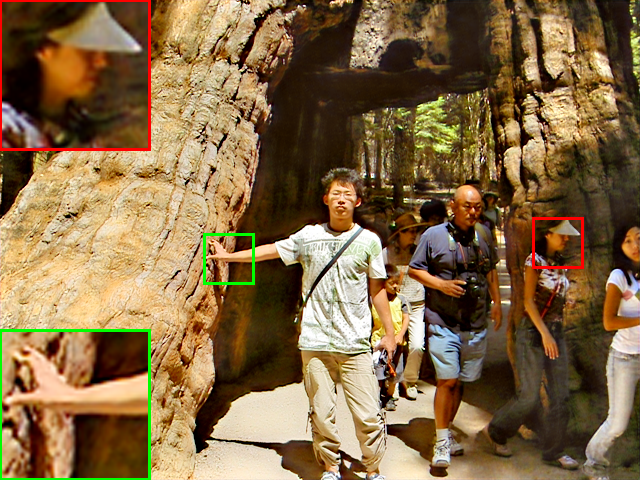} &
			\includegraphics[width=0.19\columnwidth]{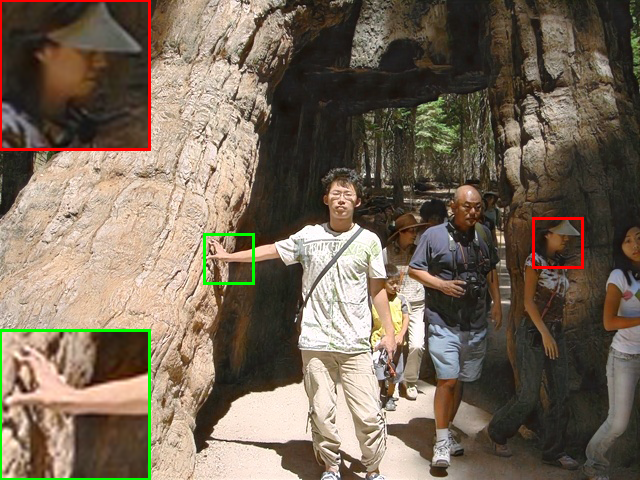} \\
			 \scriptsize{KinD} 
			 & \scriptsize{MIRNet} 
			 &\scriptsize{SSIEN} 
			 &\tiny{EnlightenGAN} 
			 & \scriptsize{DRGN (Ours)} \\
	\end{tabular}	\vspace{-2mm}
\caption{Results of different low-light image enhancement methods. RetinexNet~\cite{wei2018deep}, KinD~\cite{Zhang2019Kindling}, DeepUPE~\cite{wang2019underexposed} and EnlightenGAN~\cite{jiang2021enlightengan} light up the image contents, but cause severe color distortion. LIME~\cite{7782813}, Zero-DCE~\cite{9157813}, MIRNet~\cite{zamir2020learning} and SSIEN~\cite{zhang2020self} tend to produce over-exposed images. Our DRGN generates more realistic and credible textures with visually pleasing contrast.} 
	\vspace{-6mm}
\label{fig:sample}
\end{figure}

\noindent \textbf{Scheme One:} An observed low-light image $I_{L}$ is featured as a superposition of a normal-light image $I$ with some perturbations $I_N$ (noise, exposure, etc.) through a special combination manner $\theta(\cdot)$, defined as
\begin{equation}
\label{eq:N}
I_{L} = \theta(I, I_N).
\end{equation}
Given $I_{L}$, the goal of low-light image enhancement is to predict $I$. When $\theta(\cdot)$ is simplified as the pixel-wise summation~\cite{gharbi2017deep}, the effort is to learn the perturbation $I_N$ and then subtract it from the observed image $I_{L}$. Some methods~\cite{Lv2018MBLLEN,yang2020fidelity} also attempt to directly estimate the optimal approximation of $I$. However, these methods rely heavily on abundant data, innovative architectures, and training strategies, ignoring the intrinsic degradation itself, resulting in unsatisfactory and unnatural contents in texture, color, contrast, etc.

\noindent \textbf{Scheme Two:} The low-light image $I_{L}$ is mathematically modeled as a combination of the reflectance $I_R$ and illumination $I_{T}$ component through Eq.~\ref{eq:D2}:
\begin{equation}
\label{eq:D2}
I_{L} = I_{R}\circ I_{T},
\end{equation}
where $\circ$ denotes the element-wise product. With this kind of scheme, an image can be enhanced either by estimating and adjusting the illumination map~\cite{li2018lightennet,wang2019underexposed}, or by learning joint features of these two components~\cite{Zhang2019Kindling}. It is known as the Retinex based method~\cite{land1977retinex, wei2018deep}. These methods follow the layer decomposition paradigm and show impressive results in stretching contrast and removing noise in some cases, but still have three problems. 1) Traditional models rely on properly enforced priors and regularization. 2) Deep RetinexNets require a large number of paired training samples. 3) The enhanced reflectance $I_R$ is often treated as the normal-light image~\cite{yue2017contrast}, which often fails to produce high-fidelity and naturalness results due to the gap between the ideal situation and reality with simplified imaging models.

In sum, there are two main drawbacks of these two schemes. First, their performances \textit{heavily rely on the diversity and quality of synthetic training samples}. 
Second, they \textit{simplify the imaging model} when coping with the enhancement task, thus \textit{sacrificing representation precision}, especially when being trained on a small dataset. More specifically, since the degradation additionally damages the texture, color, and contrast of normal-light images, we argue that subtracting the perturbation $I_N$ from its low-light input $I_{L}$ directly (\textbf{scheme one}) cannot fully recover multiple kinds of fine details and color distribution (refer to MIRNet~\cite{zamir2020learning} in Figure~\ref{fig:sample}). Meanwhile, adjusting the illumination map to enhance reflectance $I_R$ (\textbf{scheme two}) cannot fully keep the visual naturalness of the predicted normal-light image
(refer to RetinexNet~\cite{wei2018deep} and DeepUPE~\cite{wang2019underexposed} in Figure~\ref{fig:sample}).

In light of the color image formulation, an image is composed of the diffuse illumination color and the environment illumination color, determined by the light source, albedo (material), environment light, noise, etc~\cite{wu2017interactive}. We thus promote \textbf{scheme one} by considering the blending relations between the normal-light image $I$ and the degradation $I_{D}$, denoted as $f(I_{D}, I)$. The newly designed low-light image generation process can be modeled as
\begin{equation}
\label{eq:D3}
I_{L} = I_{D} + I + f(I, I_{D}) =  I_{D} + \psi(I_{D}, I). 
\end{equation}
We argue ($I$, $f(I_{D}, I)$) represents the intrinsic manifold projection from $I$ to $I_L$ via degradation $I_{D}$. We formulate this complex mapping as a high-dimensional non-analytic transfer map $\psi(I, I_{D})$ since there is no explicit analytic function to present the blending relations caused by the albedo (material) ($I$), environment light, noise, etc ($I_{D}$).

By doing so, we believe the degradation $I_{D}$ can be learned from the low-light input $I_{L}$ via a degradation generator (DeG), formulated as $\phi(I_{L})$. Thus, given $I_L$, we use Eq.~\ref{eq:DeAndRe} to predict the normal-light image $I$ as
\begin{equation}
\label{eq:DeAndRe}
I = \psi^{-1}(I_{L} - \phi(I_{L})).
\end{equation} 
To alleviate the learning difficulty, our enhancement task is then decomposed into two stages: (1) simulating the degradation $I_{D}$ via DeG, and (2) refining the color and contrast information by a refinement generator (ReG), which is parameterized by $\psi^{-1}(\cdot)$ (learning the inverse representation of $\psi(\cdot)$). More specifically, inspired by the impressive results of GAN in image synthesis and translation~\cite{GatysEB15,bulat2018to}, we train a parametric generator to learn the degradation factors from the low-light input in the first stage, denoted as $\phi(I_{L})$. Note that simulated degradation factors are also applied to another referenced high-quality dataset to generate synthetic low-light samples to mitigate the limitation of sample monotonicity and repetitiveness (More details on the data generation are described in Sec.~\ref{sec:DA}). Meanwhile, we produce the base enhancement result $I_B$ by removing the predicted degradation from the low-light input. Thus, the second stage takes $I_B$ as input, and applies ReG to estimate the inverse transformation matrix to further refine the color and textural details.

DeG is implemented with a multi-resolution fusion strategy because the degradation always acts on various texture richness levels. We thus propose a multi-resolution fusion network (MFN) to learn the joint representation of multi-scale features. In the second stage, ReG shares the same architecture as DeG for simplicity, since the cooperative representation among the multi-scale features also contributes more to the recovery of image textures.
Our main contributions are summarized as follows:
\begin{itemize}
    \item We propose to decompose and characterize the low-light degradation. Thanks to the degradation generator, it can augment an arbitrary number of paired samples using a small synthetic dataset, and achieves impressive performance and robustness. Our proposed degradation formulation can be easily extended to other enhancement tasks, to help get rid of paired dataset collection. Based on this, we construct a novel two-stage Degradation-to-Refinement Generation Network (DRGN) by cascading DeG and ReG for low-light image enhancement tasks.
    \item Besides achieving impressive enhancement performance (surpassing KinD++~\cite{zhang2021beyond} by 0.70dB in PSNR on average), our method is also competitive 
        on the joint low-light image enhancement and detection tasks on ExDark~\cite{Exdark} and NightSurveillance~\cite{NightSurveillance} datasets.
    \item We also introduce two novel low/normal-light datasets (COCO24700 for training and COCO1000 for evaluation) based on COCO~\cite{caesar2018coco} dataset, which can be used for the joint low-light image enhancement and detection tasks under extreme night-time degradation scenarios. 
\end{itemize}

\begin{figure*}[!ht]
\flushleft
\centering
\includegraphics[width=0.8\textwidth]{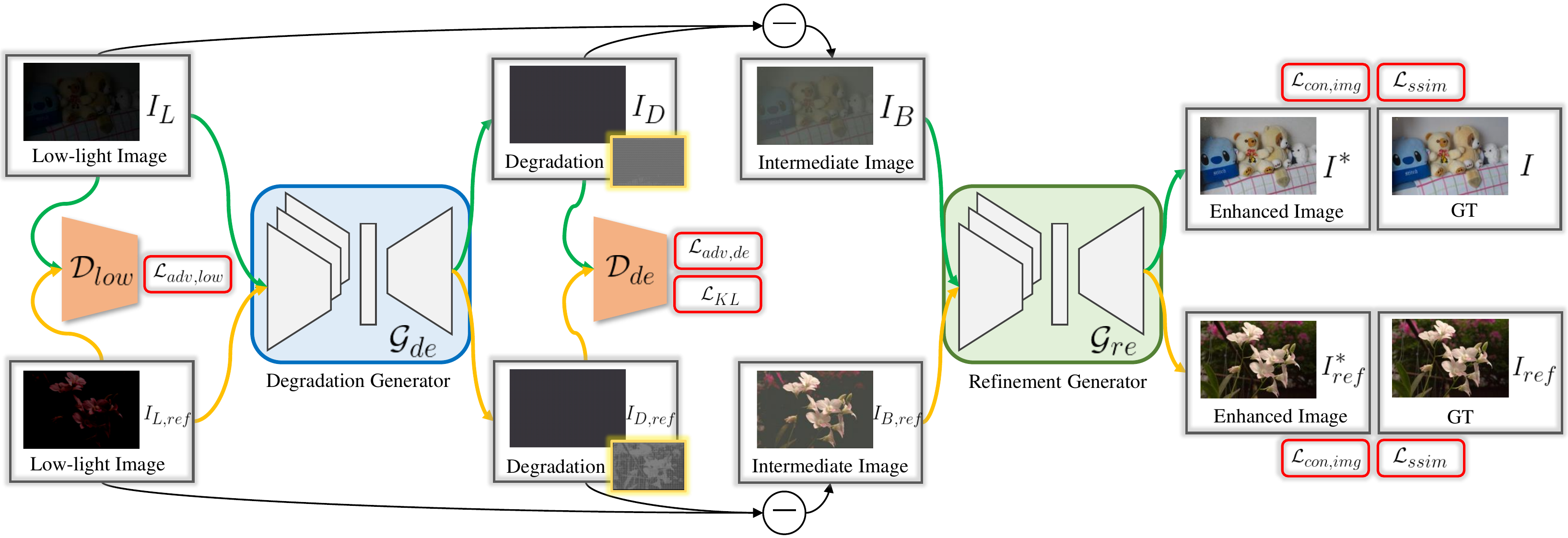}
\vspace{-3mm}
\caption{Architecture of Degradation-to-Refinement Generation Network (DRGN). It consists of two subnetworks to tackle degradation estimation and content refinement, respectively. A degradation generator (DeG) learns degradation $I_{D}$ from the low-light input $I_{L}$ in the first stage, and produces the base enhanced result $I_B$ by removing $I_{D}$. Then, a refinement generator (ReG) takes $I_B$ as input and produces the refined prediction ($I^{*}$) of the normal-light image ($I$). We also apply DeG to generate the synthetic paired samples $[I_{L,ref}, I_{ref}]$ to augment the sample space to help train these two generators.
}
\label{fig:Framework}
\vspace{-4mm}
\end{figure*}

\section{Method}
\label{sec:Me}
Figure~\ref{fig:Framework} shows the overall pipeline of our proposed low-light enhancement model -- Degradation-to-Refinement Generation Network (DRGN). The objective of DRGN is mainly two-fold: one is to train a degradation generator (DeG) $\mathcal{G}_{de} \mapsto \phi(\cdot)$ to simulate the degradation from low-light inputs. The other is to estimate the inverse transformation matrix with a refinement generator (ReG) $\mathcal{G}_{re} \mapsto \psi^{-1}(\cdot)$ to refine color and textures with the augmented dataset by applying the predicted degradation priors.

\subsection{Degradation Generator}
\label{sec:DL}
Let $I_{L}$, $I$, and $I_{ref}$ respectively denote the low-light input, normal-light image, and high-quality reference image (not in the paired training dataset). In the first stage, we learn the degradation through a generative model where all involved parameters can be inferred in a data-driven manner. Specifically, given a low-light input $I_{L}$, we train a degradation generator (DeG) to predict the degradation $I_{D}$ from $I_{L}$, \textit{e.g.}, $I_{D} = \mathcal{G}_{de}(I_{L})$. Thus we can generate the synthetic low-light image $I_{L,ref}$ by combining $I_{D}$ and $I_{ref}$. Moreover, we produce the degradation prediction $I_{D,ref}$ from $I_{L,ref}$ using the same generator DeG. More details about the data augmentation are described in Sec.~\ref{sec:DA}.

Recall that the degradation style of the input $I_{L}$ and the synthetic low-light image $I_{L,ref}$ should be shared, we regularize both the predicted degradation prior $[I_{D}, I_{D,ref}]$ and the low-light image pair $[I_{L}, I_{L,ref}]$ to be close. 
We introduce the generative adversarial loss~\cite{8099502} and degradation consistency loss (the KL Divergence)
to train both DeG and the discriminator. Here, the generative adversarial loss between the low-light input and synthetic low-light image is defined as
\begin{equation}
\label{eq:GD1}
\mathcal{L}_{adv,low} = -\log \mathcal{D}_{low}(I_{L})-\log(1-\mathcal{D}_{low}(I_{L,ref})).
\end{equation}
In addition, the constraints on the predicted degradation priors $[I_{D}, I_{D,ref}]$ are expressed as
\begin{equation}
\label{eq:GD2}
\begin{split}
\mathcal{L}_{adv,de} &= -\log \mathcal{D}_{de}(I_{D})-\log(1-\mathcal{D}_{de}(I_{D,ref})),\\
\mathcal{L}_{kl} &= \sum_{} p(I_{D})log\frac{p(I_{D})}{p(I_{D,ref})},
\end{split}
\end{equation}
where $p(\cdot)$ denotes the target distribution. By setting $\alpha$ to $10^ {-5}$~\cite{jiang2019atmfn} to balance the GAN losses and consistency loss, the total loss in the first stage is given by
\begin{equation}
\mathcal{L}_{DeG} = \alpha \times (\mathcal{L}_{adv,low} + \mathcal{L}_{adv,de}) + \mathcal{L}_{kl}.
\label{eq:loss3}
\end{equation}

\subsection{Refinement Generator}
\label{sec:Re}
We design the refinement generator (ReG) in the second stage using the same architecture of DeG for convenience. In the first stage, we also produce the base enhanced result $I_B$ by removing $I_{D}$ from the low-light input $I_L$. Thus ReG takes $I_B$ as input and learns to refine the contrast and textural details. More importantly, except for the original sample pairs [$I_B$, $I$], we also generate additional paired images [$I_{B,ref}$, $I_{ref}$] for sample augmentation in the first stage. With the enrichment and augmentation of training samples, a more robust image enhancement model can be achieved. The procedures in the second stage can be formulated as
\begin{equation}
\label{eq:stage2}
I^* = \mathcal{G}_{re}(I_B),
I^*_{ref} = \mathcal{G}_{re}(I_{B,ref}),
\end{equation}
where $I^*$ and $I^*_{ref}$ are the predicted normal-light images via ReG. Unlike the adversarial loss in the first stage, we introduce the content loss (the Charbonnier penalty function~\cite{lai2017deep,jiang2021rain}) and the SSIM~\cite{wang2004ssim} loss between the predicted normal-light image and the high-quality ground truth to guide the optimization of ReG. The loss functions are expressed as
\begin{equation}
\small
\begin{split}
\mathcal{L}_{con, img} &= \sqrt{(I^*- I)^2+\varepsilon^2} + \sqrt{(I^*_{ref}- I_{ref})^2+\varepsilon^2},\\
\mathcal{L}_{ssim} &= SSIM(I^*, I) + SSIM(I^*_{ref}, I_{ref}),\\
\mathcal{L}_{ReG} &= \mathcal{L}_{con, img} + \lambda \times \mathcal{L}_{ssim},
\end{split}
\label{eq:CEN}
\end{equation}
where $\lambda$ is used to balance the loss terms, and experimentally set as $-0.2$. The penalty coefficient $\varepsilon$ is set to $10^ {-3}$.

\subsection{Data Augmentation}
\label{sec:DA}
Previous hand-crafted methods, such as RetinexNet, provide an effective way for data synthesis, but still with limitations:
1) Misalignment errors of producing normal-light images;
2) Complex and diverse hyper-parameters and camera settings;
3) Limited and discrete sample space. Unlike them, our DRGN produces a \textit{full-automatic pipeline to create infinite paired samples with controllable degradation strengths.}

Given a synthetic paired dataset $\mathcal{S} = \{I, I_L\}$ and a high-quality reference dataset $\mathcal{R} = \{I_{ref}\}$, we first train a generator to learn the degradation $I_{D}$ from each $I_{L}$ in the first stage. According to Eq.~\ref{eq:D3}, the ideal low-light image can be produced via $I_{L}$ = $I_{D}$ + $\psi(I, I_{D})$. In theory, there exits a \textbf{strictly high-quality} version of $I_{ref}$ to fit Eq.~\ref{eq:D3}, formulated as $I_{L,ref}$ = $I_{D}$ + $\psi(I^*_{ref}, I_{D})$, where $I^*_{ref}$ denotes the normal-light version of $I_{ref}$, and $\psi(I^*_{ref}, I_{D})$ is equal to $I_{ref}$, generated via $\psi(\cdot)$ and $I_{D}$. Since low-light image $I_{L,ref}$ and its high-quality version are not one-to-one correspondence to each other, we thus adjust the degradation mapping of $I_{L,ref}$. As shown in Figure~\ref{fig:Framework},  $I_{L,ref}$ is reformulated as $I_{L,ref}$ = $I_{D,ref}$ + $\psi(I_{ref}, I_{D,ref})$, which strictly fits the definition in Eq.~\ref{eq:D3}. To maintain the diversity and reality of the degradation, we introduce GAN loss and consistency loss to regularize both predicted degradation $[I_{D}, I_{D,ref}]$ and low-light image pair $[I_{L}, I_{L,ref}]$ to have similar distributions (\textit{Fitting results of Y channel histogram are provided in Supplementary.}).

Overall, the aforementioned procedure leads to a new paired dataset $\mathcal{T} = \{I_{ref}, I_{L,ref}\}$ during training. Therefore, our sample space consists of the original sample dataset $\mathcal{R}$ and the new synthetic dataset $\mathcal{T}$. Our degradation learning and data augmentation scheme is more like a transfer learning of degradation patterns from the normal-light sample domain to the low-light image domain. To better understand the degradation learning in our proposed study, we show examples of the predicted degradation and the synthetic low-light images in the \textit{Supplementary Material}.

\begin{figure}[t]
\flushleft
\centering
\includegraphics[width=0.95\columnwidth]{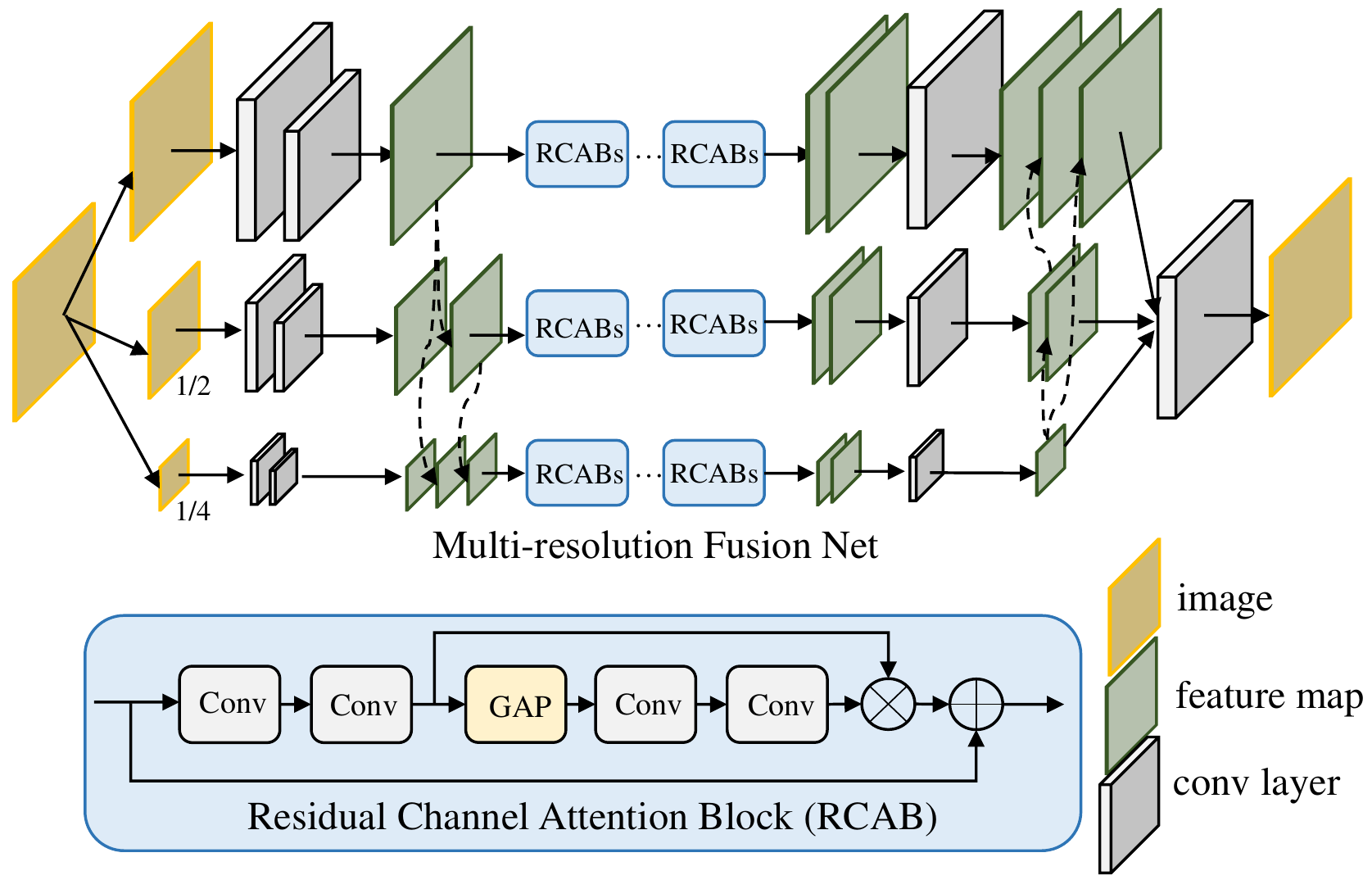}
\vspace{-2mm}
\caption{Pipeline of multi-resolution fusion network.}
\label{fig:mfn}\vspace{-6mm}
\end{figure}

\subsection{Multi-resolution Fusion Network}
Decomposing a complex task into several sub-problems and further learning their cooperative representation to yield a final solution is a practical scheme to promote the model performance~\cite{7327182,8767931}. To this end, we propose to decompose the learning task (including degradation learning and content refinement) into multiple sub-spaces, and construct a multi-resolution fusion network (MFN) to represent the target information in a multi-scale collaborative manner, as demonstrated in Figure~\ref{fig:mfn}.

Taking the procedure of degradation learning in DeG for instance, we first generate the image pyramid from the low-light input via Gaussian sampling kernel $G(\cdot)$, and adopt initial convolutions~\cite{zhang2018image} to project the input into the feature space to extract initial features $f_{ini}^{i}$, expressed as
\begin{equation}
\label{eq:ini}
f_{ini}^{i} = \mathcal{H}_{ini}^{i}(G(I_{L})), i\subseteq [1,n],
\end{equation}
where $n$ and $i$ denote the sampling number and the corresponding $i_{th}$ pyramid layer. In particular, the Gaussian sampling is activated except $i=1$. Before passing $f_{ini}^{i}$ into the backbone network, we sample the features of high-level branches with strided convolutions, and aggregate them with the features of lower-level branches to form the multi-scale representation (MSR), formulated as
\begin{equation}
\label{eq:ini2}
f_{msr}^{i} = \mathcal{H}_{msr}^{i}(f_{ini}^{i}, f_{ini}^{j}), j\subseteq [1,i),
\end{equation}
where $\mathcal{H}_{msr}^{i}(\cdot)$ denotes $i_{th}$-layer multi-scale fusion. After that, we apply the residual group $\mathcal{H}_{RG}^{i}(\cdot)$ composed of several cascaded residual channel attention blocks (RCABs), each containing multiple RCABs to build deep feature representations~\cite{zhang2018image}, as shown in Figure~\ref{fig:mfn}. A $1\times 1$ convolution is followed to perform the long-term fusion among the output and input of the residual group. The procedures above can be expressed as
\begin{equation}
\label{eq:ini3}
f_{long}^{i} = \mathcal{H}_{1}^{i}([\mathcal{H}_{RG}^{i}(f_{msr}^{i}), f_{msr}^{i}]).
\end{equation}

Through the $n$-layer pyramid network, we obtain $n$ outputs. To better characterize the degradation, we utilize the deconvolution layer to up-sample the outputs of low-level pyramid layer, and then perform the multi-scale fusion with high-level features via a $1\times 1$ convolution. Consequently, a reconstruction layer $\mathcal{H}_{rec}(\cdot)$ with the filter size of $3\times 3$ is used to project the output of MFN back to the image space. The overall procedure can be summarized as
\begin{equation}
\label{eq:ini4}
I_{D} =  \mathcal{H}_{rec}(\mathcal{H}_{msr}^{j}(f_{long}^{j}, f_{long}^{1})), j\subseteq [2,n].
\end{equation}

\section{Experiments}
\label{sec:Ex}
We carry out extensive experiments on synthetic and real-world low-light image datasets to evaluate 
our proposed DRGN. Nine representative low-light enhancement algorithms are compared, including 
LIME~\cite{7782813}, MIRNet~\cite{zamir2020learning}, KinD~\cite{Zhang2019Kindling}, KinD++~\cite{zhang2021beyond}, DeepUPE~\cite{wang2019underexposed}, RetinexNet~\cite{wei2018deep}, Zero-DCE~\cite{9157813}, SSIEN~\cite{zhang2020self} and EnlightenGAN~\cite{jiang2021enlightengan}. Peak Signal to Noise Ratio (PSNR), Structural Similarity (SSIM) and Feature Similarity (FSIM)~\cite{zhang2011fsim} are employed for performance evaluation on synthetic datasets. We also conduct additional experiments on real-world low-light datasets and adopt three no-reference quality metrics (Naturalness Image Quality Evaluator (NIQE)~\cite{mittal2012making}, Spatial-Spectral Entropy-based Quality (SSEQ) index~\cite{liu2014no} and user study scores) for evaluation.

\subsection{Implementation Details}
\label{sec:Id}
\noindent \textbf{Data Collection.} Following the setting in RetinexNet~\cite{wei2018deep}, we use the LOL dataset for training, which contains 500 low/normal-light image pairs (480 for training and 20 for evaluation). The competing methods are also trained on LOL using their publicly released codes for a fair comparison. Similar to~\cite{wei2018deep,Zhang2019Kindling}, we evaluate our model and other methods on three widely used synthetic low-light datasets: \textit{LOL1000}~\cite{wei2018deep}, \textit{Test148}~\cite{jiang2021enlightengan} and \textit{VOC144}~\cite{Lv2018MBLLEN}. Besides, we also collect 63 real-world low-light samples from~\cite{lee2012contrast,ma2015perceptual}, namely  \textit{Real63}, for evaluation. Finally, a synthetic low-light dataset (COCO1000) and a real-world low-light dataset (ExDark~\cite{Exdark}) are used to verify the performance on both image enhancement and detection tasks under low-light conditions.

\noindent \textbf{Experimental Setup.} In our baseline, the pyramid layer is empirically set to 3, corresponding to the number of RCABs and RCAB depths of [2, 3, 4] and [3, 3, 3], respectively, in the residual group. The training images are cropped to non-overlapping $96\times96$ patches to obtain sample pairs. Standard augmentation strategies, \textit{e.g.}, scaling and horizontal flipping are applied. We use the Adam optimizer with a batch size of 16 for training DRGN on a single NVIDIA Titan Xp GPU. The learning rate is initialized to $5\times10^{-4}$ and then attenuated by 0.9 every 6,000 steps. After 60 epochs on training datasets, we obtain the optimal solution with the above settings. Specifically, we train DeG for the first 20 epochs and then optimize ReG for 40 epochs.

\tabcolsep=3pt
\begin{table}[t]
  \centering
  \scriptsize
  \begin{tabular}{cccccccc}
  \hline
  Model & DL-DA &  MSR & $\mathcal{L}_{KL}$ & $\mathcal{L}_{ssim}$ &  PSNR &  SSIM & FSIM \\
  \hline
  Input   & --  & -- & --    & -- & 12.86& 0.610& 0.787\\
  \textit{w/o} DL-DA   & $\times$   & $\checkmark$  & $\times$ & $\checkmark$ & 19.03& 0.857& 0.906\\
  \textit{w/o} MSR   & $\checkmark$ & $\times$   & $\checkmark$   & $\checkmark$ & 19.32& 0.874& 0.910\\
  \textit{w/o} KL  & $\checkmark$ & $\checkmark$ & $\times$  & $\checkmark$ & 19.61& 0.881& 0.914\\
  \textit{w/o} SSIM  & $\checkmark$ & $\checkmark$  & $\checkmark$ & $\times$ & 19.56& 0.849& 0.907\\
  DRGN & $\checkmark$ & $\checkmark$ & $\checkmark$  & $\checkmark$& \textbf{19.88}& \textbf{0.889}& \textbf{0.916}\\
  \hline
  \end{tabular}
  \caption{Ablation study of basic components on LOL1000 dataset. DL-DA, MSR, KL and SSIM stand for the degradation learning and data augmentation, multi-scale representation, KL divergence and SSIM constraint, respectively.}
  \label{table:ablation}
\end{table}

\begin{figure}[!t]
	\centering
	\includegraphics[width=0.95\linewidth]{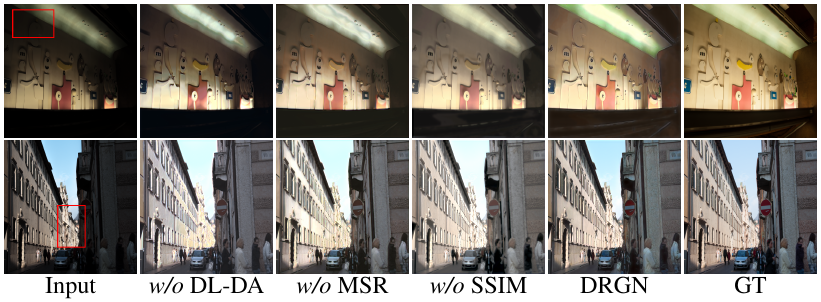}
	\vspace{-2mm}
\caption{Visualization results 
on LOL1000 dataset.} 
\label{fig:ablation}\vspace{-6mm}
\end{figure}

\begin{table*}[t]
  \centering
  \scriptsize
  \tabcolsep=2.5pt
  \resizebox{1\linewidth}{!}{
  \begin{tabular}{cccccccccccc}
  \toprule
  Methods & LIME 
  & RetinexNet
  & DeepUPE$^*$
  & Zero-DCE
  &KinD
  &KinD++ 
  &MIRNet
  &SSIEN
  &EnlightenGAN
  & DRGN (Ours)& LW-DRGN (Ours)\\
  \midrule
  PSNR~$\uparrow$ & 18.35/18.32/19.38 & 15.45/18.23/18.08 & 18.41/16.64/19.43 & 14.42/15.53/15.42 & 17.27/18.93/22.19 & 19.18/\underline{19.42/22.45}
  &\underline{20.34}/17.20/19.13 & 16.23/17.35/18.18 &16.34/17.64/19.49
  &\textbf{20.41/19.88/22.86} & 18.85/18.89/20.43\\
  SSIM~$\uparrow$  & 0.771/0.829/0.757& 0.758/0.792/0.749 & 0.766/0.773/0.781 & 0.392/0.420/0.348 &0.832/0.851/0.834 & 0.859/\underline{0.857/0.842}
  &\underline{0.871}/0.813/0.781&0.780/0.779/0.727 &0.796/0.835/0.790
  &\textbf{0.880/0.889/0.858}& 0.851/0.856/0.839\\
  FSIM~$\uparrow$  & 0.896/0.871/0.827& 0.868/0.860/0.837 & 0.910/0.865/0.873 & 0.738/0.741/0.635 &0.900/0.908/0.881 & 0.923/\underline{0.911/0.885}
  &\underline{0.928}/0.872/0.850&0.870/0.830/0.792 &0.907/0.884/0.853
  &\textbf{0.940/0.916/0.894} & 0.920/0.886/0.882\\
  Average~$\uparrow$ & 18.68/0.785/0.864 & 17.25/0.766/0.855 & 18.16/0.773/0.882 &15.12/0.386/0.704 &19.46/0.839/0.896 &\underline{20.35/0.852/0.906}
  &18.89/0.821/0.883  & 17.25/0.762/0.830  &17.82/0.807/0.881  &\textbf{21.05/0.875/0.916}  & 19.39/0.848/0.896\\
  \midrule
  Par.(M)~$\downarrow$  & --& 0.445 & 0.999& \textbf{0.079} & 8.016 & 8.274 &31.78&\underline{0.486} &8.64& 4.773& 0.660\\
  Time (S)~$\downarrow$ & --& 0.414 & \underline{0.016} & \textbf{0.010} & 0.059 & 0.392 &0.630&0.028 &0.057 &0.152& 0.041\\
  GFlops (G)~$\downarrow$ & --& 83.57 &  \textbf{0.214} & \underline{11.74} & 78.40 & 1670.91 &554.32&77.86&37.02 &110.59& 18.15\\
  \bottomrule
  \end{tabular}}\vspace{-3mm}
  \caption{Quantitative performance comparison of the compared methods on Test148/LOL1000/VOC144. LW-DRGN denotes our light-weight model. We also show the number of model parameters (Million), average inference time (Second) and GFlops used for 384 $\times$ 384 images, except for the traditional prior-based algorithm LIME. $^*$ denotes that results are obtained with the released test codes directly. The \textbf{bold font} and \underline{underline} represent the best and second performances, respectively.} 
  \label{table:synthetic}
\end{table*}

\begin{figure*}[t]
	\centering
	\includegraphics[width=0.95\textwidth]{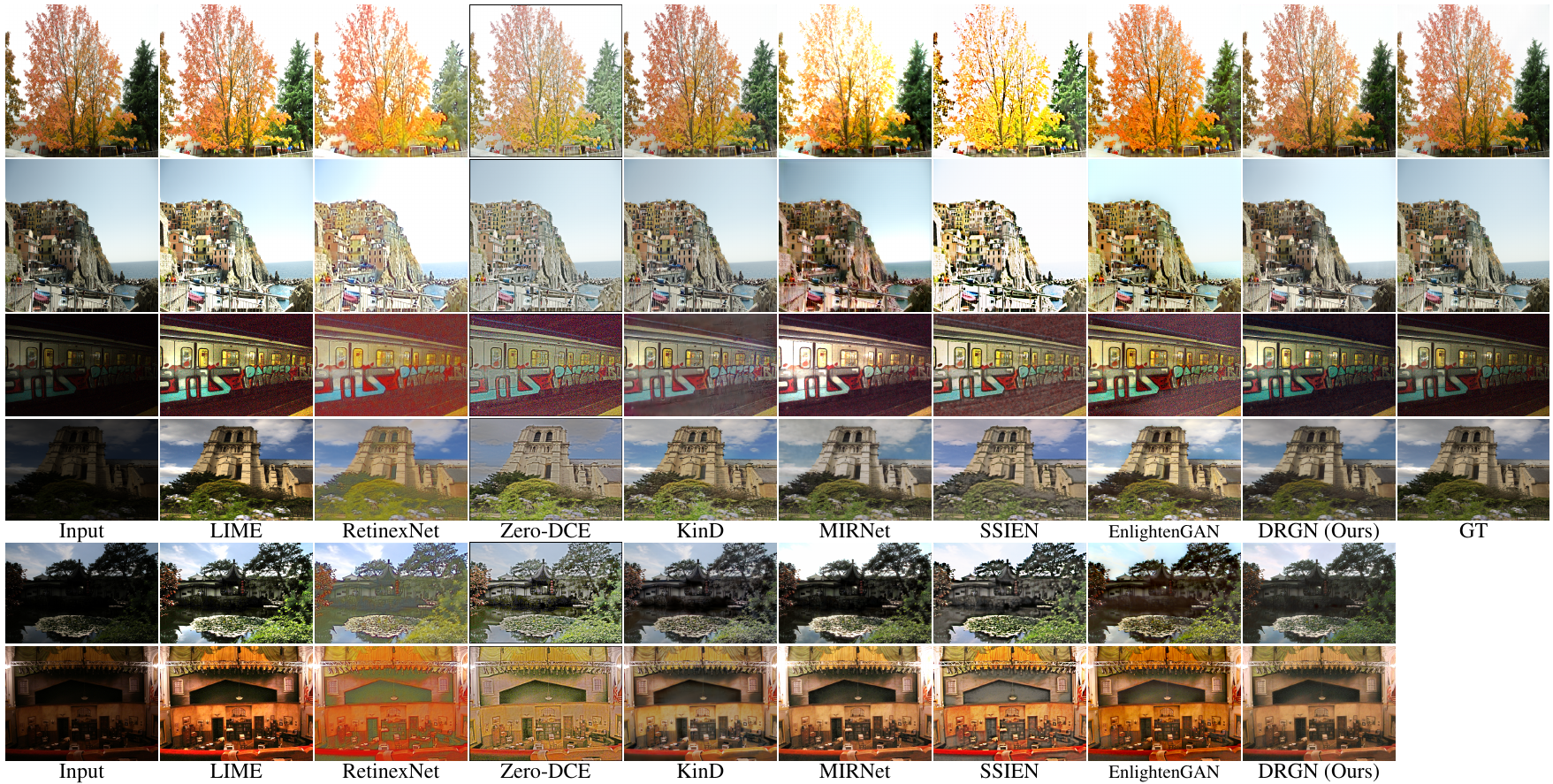}
	\vspace{-3mm}
   \caption{Restoration results on synthetic ($1^{st}-4^{th}$ rows) and and real-world scenarios ($5^{th}-6^{th}$ rows).}
\label{fig:synthetic}\vspace{-6mm}
\end{figure*}

\subsection{Ablation Study}
\label{sec:As}
\noindent \textbf{Basic Components.}~Since the proposed DRGN involves multiple enhancement strategies, including degradation learning and data augmentation (DL-DA) and multi-scale representation (MSR), we carry out ablation studies to validate the contributions of individual components to the final enhancement performance. We design two models by removing these two components in turn while keeping the similar computational complexity, termed \textit{w/o} DL-DA and \textit{w/o} MSR. Quantitative results on the LOL1000 dataset are tabulated in Table~{\ref{table:ablation}}, showing that DRGN containing all modules significantly outperforms its incomplete versions. For example, with the degradation learning and data augmentation, DRGN gains about 0.85dB performance improvements over \textit{w/o} DL-DA.  We consider that applying a parametric generator to characterize the implicit distribution of degradation is more reasonable and practical, avoiding the dependence on hand-crafted priors. Furthermore, the training dataset can be automatically augmented and enriched with the newly generated synthetic paired samples, which promote the diversity and abundance of the training dataset. On the other hand, using the multi-scale representation strategy can cope with the learning task in multiple subspaces, thus effectively alleviating the learning difficulty with the joint representation among multi-scale features. The results in Figure~\ref{fig:ablation} also demonstrate that the proposed modules can together produce results with more credible contents and faithful contrast.

\noindent \textbf{Loss Function.} We also evaluate the influence of KL divergence and SSIM loss on the enhancement performance by training another two models, \textit{w/o} KL (optimized without KL divergence in DeG)  and \textit{w/o} SSIM (optimized without SSIM loss in ReG), for comparison. Comparison results on LOL1000 are shown in Table~{\ref{table:ablation}} and Figure~\ref{fig:ablation}. KL divergence is used to constrain DeG to keep the distribution consistency between the predicted degradation priors [$I_D$, $I_{D,resf}$] and accelerate the training. SSIM loss can guide the model to produce results with better texture fidelity by enforcing structural similarity. Therefore, compared with \textit{w/o} KL or \textit{w/o} SSIM models, DRGN gains notable improvements in quality metrics and visual effects.
\subsection{Comparison with State-of-the-Arts}
\noindent \textbf{Synthetic Data.} Nine representative low-light image enhancement methods are employed for comparison. Quantitative results on LOL1000~\cite{wei2018deep}, Test148~\cite{jiang2021enlightengan} and VOC144~\cite{Lv2018MBLLEN} datasets are presented in Table~{\ref{table:synthetic}}. Our DRGN significantly outperforms the competing methods among all quality metrics. In particular, DRGN outperforms the current SOTA approaches -- KinD++~\cite{zhang2021beyond} and EnlightenGAN~\cite{jiang2021enlightengan} -- \textbf{in PSNR (SSIM) by 0.70dB (0.023) and 3.23dB (0.068) on average while enjoying 42.3\% and 93.3\% more efficiency on model parameters and computation cost.} These substantial improvements validate the efficacy of the proposed modules (degradation learning and data augmentation, and multi-scale fusion representation) for the accurate recovery of image content and contrast.

Efficiency also plays an important role in evaluating the model performance for real-time applications in particular. To this end, we construct a light-weight model, namely LW-DRGN. Specifically, we simplify the multi-resolution fusion network with the single-branch residual network to reduce the parameter size and the memory footprint. Meanwhile, an attention-guided dense fusion strategy is also used to replace the original skip connection between modules to promote the discriminative representation of multi-stage features. Quantitative results in Table~{\ref{table:synthetic}} show that LW-DRGN yields an acceptable performance (0.07dB less than KinD on average), but \textbf{only requires 8.37\%, 23.2\% and 69.5\% in parameters, computation complexity (Gflops) and inference time.}
The high efficiency and effectiveness of LW-DRGN makes it practical for mobile devices and applications requiring real-time throughput.

\begin{figure}[!ht]
	\centering
	\renewcommand\arraystretch{0.3}
		\begin{tabular}{cc}
			\includegraphics[width=0.49\columnwidth]{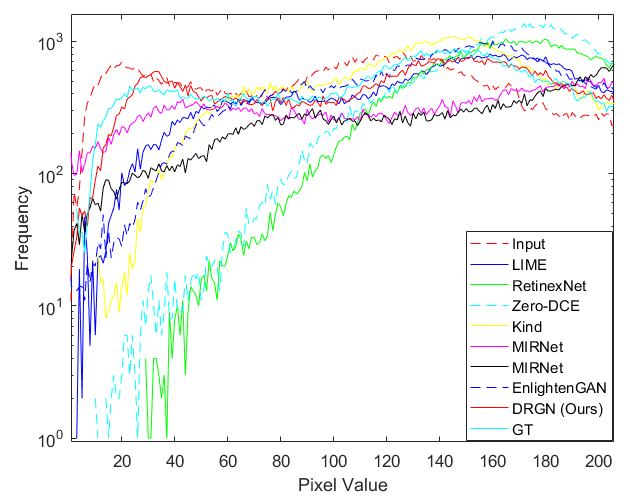} &
			\includegraphics[width=0.49\columnwidth]{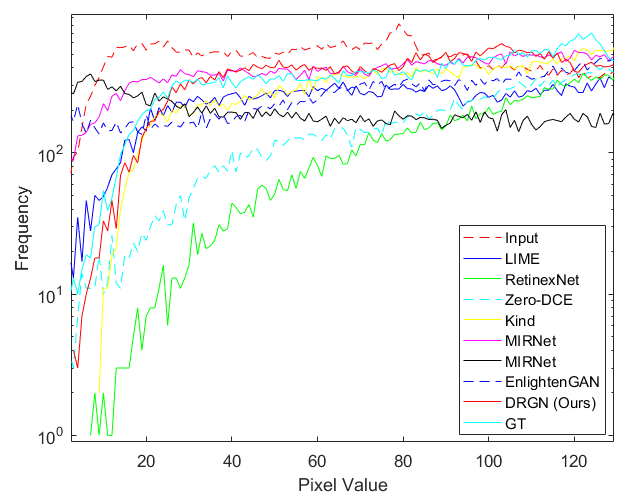} \\
			\scriptsize{Tree} & \scriptsize{Building}\\
			\includegraphics[width=0.49\columnwidth]{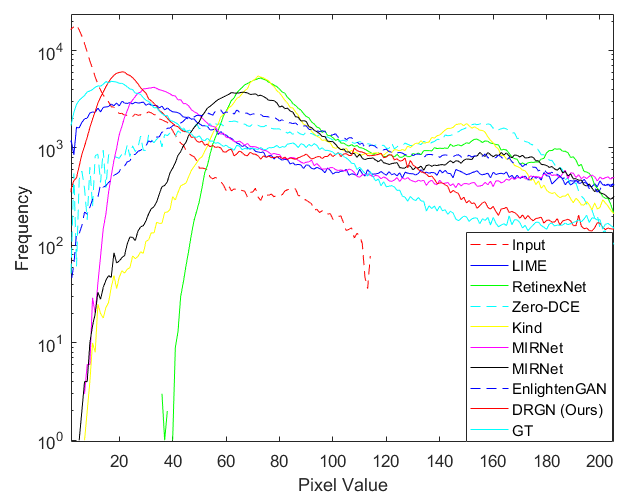} &
			\includegraphics[width=0.49\columnwidth]{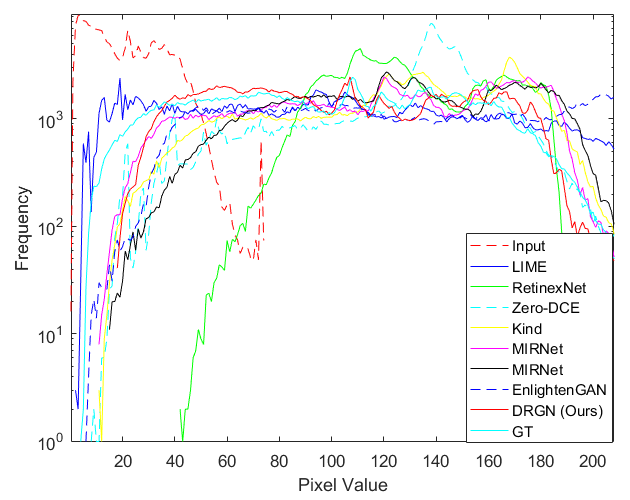} \\
			\scriptsize{Train}& \scriptsize{House}
	\end{tabular}
   \caption{The fitting results based on the histogram curve of Y channel in YCbCr on four synthetic scenarios in Figure~\ref{fig:synthetic}.}
\label{fig:fitting}\vspace{-2mm}
\end{figure}

Qualitative comparisons under different low-light conditions (diverse exposures and degradation) on three synthetic low-light datasets are depicted in Figure~\ref{fig:synthetic}. The results show that DRGN can cope with all low-light cases, generating results with clear and credible details, as well as better fidelity and naturalness. Although other competitors can light-up contents in the dark regions or produce vivid visual effects (LIME and EnlightenGAN), the predicted results are still corrupted by halo artifacts and noises, and suffer from color distortion as well. For example, as observed from the ``Train" and ``House" images in Figure~\ref{fig:synthetic}, only DRGN restores credible image details and approximates the contrast more similar to the ground-truth, while the competitors fail to solve the degradation and their results exhibit apparent  color deviation. Moreover, we observe an interesting phenomenon from these results. The competitors tend to produce brighter outputs no matter what level of degradation (slight or severe degradation conditions), since they  complete the relighting task using a unified enhancement module (see the first (slight degradation) and third (severe degradation) scenarios in Figure~\ref{fig:synthetic}). However, DRGN can cope with different degradation conditions and produces results with better naturalness and texture fidelity. This is because DRGN takes advantage of the intrinsic degradation priors of the input and employs the data augmentation to improve the training. Moreover, we also provide the fitting curve based on the histogram of ``Y" channel on four synthetic scenarios in Figure~\ref{fig:fitting}, showing that our results are close to the distribution of ground truth.

\tabcolsep=0.5pt
\noindent \textbf{Real-world Data.}
Although recent years have witnessed significant progress on the low-light image enhancement task, recovering credible contents and visually pleasing contrast from real-world low-light scenarios is still challenging. Inspired by~\cite{lee2012contrast,ma2015perceptual}, we conduct experiments on the Real63 dataset to further verify the effectiveness of DRGN. Besides two reference-free metrics (NIQE and SSEQ), we also perform a user study to evaluate the performance quantitatively. Specifically, a total of 30 volunteers independently score the visual quality of the enhanced images in the range of 1.0 to 5.0 (worst to best quality) in terms of the content naturalness (CN) and color deviation (CD). The quantitative results are included in Table~{\ref{table:real}}. Again, our proposed DRGN achieves the best and second-best average scores of NIQE and SSEQ. Moreover, DRGN yields the highest average user study scores, including CN and CD, for a total of 63 scenarios. Figure~\ref{fig:synthetic} shows the visual comparison results. DRGN can infer more realistic and credible image details in dark regions and well adjust the image contrast, whereas other methods tend to produce under- or over-exposure results with noticeable color distortion (please refer to the ``sky" and ``plant" in the $5^{th}$ scenario.). These results confirm that applying degradation learning and data augmentation can promote robustness and generality in real-world scenarios.

\begin{table}[!ht]
\begin{center}
\tabcolsep=5pt
\resizebox{1\linewidth}{!}{
\begin{tabular}{lcccccccc}
\hline
Methods &LIME&\small{RetinexNet}& Zero-DCE &KinD& MIRNet& SSIEN& \small{EnlightenGAN}& \small{DRGN (Ours)}\\
\hline
  NIQE~$\downarrow$  & \underline{3.117} & 5.192 & 3.257 & 3.551 & 3.287& 4.153& 4.408& \textbf{3.051}\\
  SSEQ~$\downarrow$  & 24.45& 36.88& 23.61 & 29.48 & 26.69& 32.20& \textbf{20.08}& \underline{23.53}\\
  CN~$\uparrow$  & 3.587& 2.245 & 3.147 & 3.458& 3.569 &3.026 & \underline{3.852}& \textbf{4.012}\\
  CD~$\uparrow$ & 3.531& 2.263 & 3.259 & \underline{3.762} & 3.625&3.148 & 3.471& \textbf{3.984}\\
\hline
\end{tabular}}\vspace{-1mm}
\caption{Comparison results 
on Real63 dataset. CN and CD are the user study scores. The \textbf{bold font} and \underline{underline} represent the best and second performances, respectively.}
\label{table:real}\vspace{-6mm}
\end{center}
\end{table}

\subsection{Evaluation via Downstream Vision Tasks}
\label{sec:EDVT}

\begin{figure}[!t]
	\centering
	\includegraphics[width=0.95\linewidth]{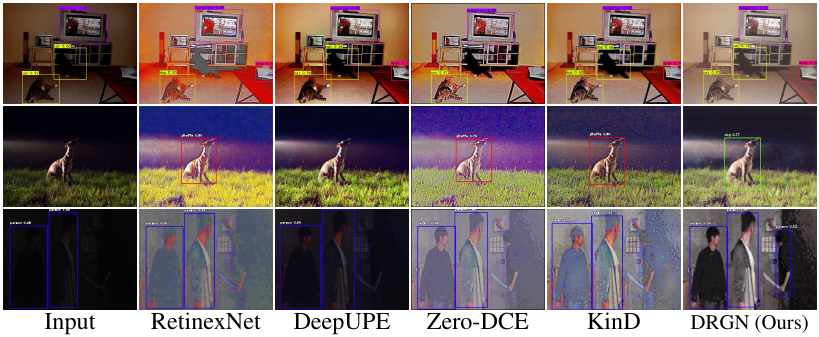}
	\vspace{-2mm}
\caption{Examples of joint low-light image enhancement and object detection on COCO1000 and ExDark.}
\label{fig:Odetection}\vspace{-6mm}
\end{figure}

Lighting up the image under low-light conditions while recovering credible textural details is meaningful for many high-level vision applications such as object and pedestrian detection. This motivates us to investigate the impact of image enhancement performance on the accuracy of object detection. Therefore, we conduct experiments for the joint image enhancement and object detection tasks. We adopt the real-world low-light dataset (ExDark~\cite{Exdark}) and the synthetic low-light dataset (COCO1000) for evaluation, where the samples are of diverse exposures and degradation and most of them are from night scenes. Our proposed DRGN, KinD~\cite{Zhang2019Kindling},  Zero-DCE~\cite{9157813}, DeepUPE~\cite{wang2019underexposed}, and RetinexNet~\cite{wei2018deep} are directly applied to the low-light samples to generate the predicted normal-light ones. We then use the pre-trained YOLOv4~\cite{yolov4} detector for object detection. Qualitative results, including the image enhancement performance and the precision of object detection, are tabulated in Table~{\ref{table:detection}}~(a) and (b). Both the enhancement performance and detection precision on predicted normal-light images by DRGN show a significant improvement over the competing low-light enhancement methods, surpassing the top-performing method KinD by $1.91dB$ and $2.42\%$ in terms of PSNR and mAP on COCO1000 dataset. DRGN also achieves the best results of $63.83\%$ IoU and $67.72\%$ precision on ExDark. Visual comparisons in Figure~\ref{fig:Odetection} clearly demonstrate that low-light degradation can greatly affect the readability and recognition, which hinders the object detection accuracy. The enhanced images by our proposed DRGN method obviously promote the detection precision.

\tabcolsep=3pt
\begin{table}[!ht]\vspace{-2mm}
\begin{center}
\resizebox{\linewidth}{!}{
\begin{tabular}{ccccccc}
\hline
& Methods & Input& RetinexNet & Zero-DCE &KinD & DRGN\\
\hline
\multirow{2}*{Enhancement}  & PSNR~$\uparrow$ &11.58 &18.32 &  15.07 & 18.56 & \textbf{20.47}\\
                            & SSIM~$\uparrow$ &0.586 &0.723 & 0.368& 0.748& \textbf{0.790}\\

\hline
\multirow{3}*{Detection}  & Pre. ($\%$) & --& 25.18  & 25.72&  26.54 & \textbf{26.81}\\
           & IoU ($\%$)  &  -- &76.54&  76.19 &  77.01& \textbf{77.35}\\
           & mAP ($\%$)  &  --& 37.58&  41.85 &  43.49& \textbf{45.91}\\
\hline
\end{tabular}
}

(a) COCO1000


\resizebox{\linewidth}{!}{
\begin{tabular}{clcccccc}
\hline
& Methods & Input & RetinexNet & Zero-DCE &KinD & DRGN\\
\hline
\multirow{2}*{Enhancement} & NIQE~$\downarrow$ & 4.887&5.225 &  3.825 & 3.770 & \textbf{3.712}\\
  & SSEQ~$\downarrow$ &31.58&36.98  & 34.27 & 36.04& \textbf{30.65}\\
\hline
\multirow{3}*{Detection}  & Pre. ($\%$) & 63.95& 62.84  &  63.91 &  64.86 & \textbf{67.72}\\
  &  IoU ($\%$) &  53.28 & 56.17&  58.48&  59.29& \textbf{63.83}\\
  & mAP ($\%$)  &  69.34 & 67.72&  68.21 &  71.45& \textbf{74.63}\\
\hline
\end{tabular}}
(b) ExDark~\cite{Exdark}



\caption{Comparison results of joint low-light image enhancement and object 
detection. 
Image Size: 640$\times$ 480; Algorithm: YOLOv4. 
We report the results on the original images and enhanced images with our DRGN.}\vspace{-2mm}
\label{table:detection}
\end{center}
\end{table}\vspace{-6mm}

\section{Conclusion}
\label{sec:Con}
We proposed a novel Degradation-to-Refinement Generation Network (DRGN) for low-light image enhancement. To achieve high-naturalness recovery, we decompose the enhancement task into two stages: degradation learning and content enhancement. Using the degradation learning, a number of low/normal-light image pairs can be generated to augment the sample space to promote the training. Moreover, the multi-resolution fusion strategy further improves the feature representation. Consequently, the recovered normal-light images are more realistic with high naturalness due to the degradation learning, data augmentation and multi-scale feature fusion. In particular, we create two low/normal-light datasets (COCO24700 for training and COCO1000 for evaluation) for the joint low-light image enhancement and objective detection tasks. Extensive experimental results on synthetic and real-world low-light samples and downstream vision tasks (object and pedestrian detection) demonstrate the superiority of DRGN over the competing methods.

\section*{Acknowledgments}
This work is supported by National Natural Science Foundation of China (U1903214, 62071339, 61872277, 62072347, 62171325), and Guangdong-Macao Joint Innovation Project (2021A0505080008).

\bibliography{aaai22}

\begin{thebibliography}{46}
\providecommand{\natexlab}[1]{#1}

\bibitem[{Bae(2019)}]{Bae2019aaai}
Bae, S. 2019.
\newblock Object Detection Based on Region Decomposition and Assembly.
\newblock In \emph{AAAI}, 8094--8101.

\bibitem[{Bochkovskiy, Wang, and Liao(2020)}]{yolov4}
Bochkovskiy, A.; Wang, C.-Y.; and Liao, H.-Y.~M. 2020.
\newblock Yolov4: Optimal speed and accuracy of object detection.
\newblock \emph{arXiv e-prints}, 2004.10934.

\bibitem[{Bulat, Yang, and Tzimiropoulos(2018)}]{bulat2018to}
Bulat, A.; Yang, J.; and Tzimiropoulos, G. 2018.
\newblock To Learn Image Super-Resolution, Use a GAN to Learn How to Do Image
  Degradation First.
\newblock In \emph{ECCV}, 187--202.

\bibitem[{Caesar, Uijlings, and Ferrari(2018)}]{caesar2018coco}
Caesar, H.; Uijlings, J.; and Ferrari, V. 2018.
\newblock Coco-stuff: Thing and stuff classes in context.
\newblock In \emph{CVPR}, 1209--1218.

\bibitem[{Chen et~al.(2018)Chen, Chen, Xu, and Koltun}]{chen2018learning}
Chen, C.; Chen, Q.; Xu, J.; and Koltun, V. 2018.
\newblock Learning to see in the dark.
\newblock In \emph{CVPR}, 3291--3300.

\bibitem[{{Fu} et~al.(2020){Fu}, {Liang}, {Huang}, {Ding}, and
  {Paisley}}]{8767931}
{Fu}, X.; {Liang}, B.; {Huang}, Y.; {Ding}, X.; and {Paisley}, J. 2020.
\newblock Lightweight Pyramid Networks for Image Deraining.
\newblock \emph{IEEE Transactions on Neural Networks and Learning Systems},
  31(6): 1794--1807.

\bibitem[{Gatys, Ecker, and Bethge(2015)}]{GatysEB15}
Gatys, L.~A.; Ecker, A.~S.; and Bethge, M. 2015.
\newblock Texture synthesis and the controlled generation of natural stimuli
  using convolutional neural networks.
\newblock In \emph{Bernstein Conference 2015}, 219--219.

\bibitem[{Gharbi et~al.(2017)Gharbi, Chen, Barron, Hasinoff, and
  Durand}]{gharbi2017deep}
Gharbi, M.; Chen, J.; Barron, J.~T.; Hasinoff, S.~W.; and Durand, F. 2017.
\newblock Deep bilateral learning for real-time image enhancement.
\newblock \emph{ACM TOG}, 36(4): 1--12.

\bibitem[{{Guo} et~al.(2020){Guo}, {Li}, {Guo}, {Loy}, {Hou}, {Kwong}, and
  {Cong}}]{9157813}
{Guo}, C.; {Li}, C.; {Guo}, J.; {Loy}, C.~C.; {Hou}, J.; {Kwong}, S.; and
  {Cong}, R. 2020.
\newblock Zero-Reference Deep Curve Estimation for Low-Light Image Enhancement.
\newblock In \emph{CVPR}, 1777--1786.

\bibitem[{{Guo}, {Li}, and {Ling}(2017)}]{7782813}
{Guo}, X.; {Li}, Y.; and {Ling}, H. 2017.
\newblock LIME: Low-Light Image Enhancement via Illumination Map Estimation.
\newblock \emph{IEEE Trans. Image Process.}, 26(2): 982--993.

\bibitem[{Huang et~al.(2021)Huang, Hu, Wang, Liang, and
  Chen}]{huang2021occluded}
Huang, W.; Hu, R.; Wang, X.; Liang, C.; and Chen, J. 2021.
\newblock Occluded suspect search via channel-guided mechanism.
\newblock \emph{Neural Comput. Appl.}, 33(3): 961--971.

\bibitem[{Huang et~al.(2018)Huang, Liang, Yu, Wang, Ruan, and
  Hu}]{HuangLYWRH2018AAAI}
Huang, W.; Liang, C.; Yu, Y.; Wang, Z.; Ruan, W.; and Hu, R. 2018.
\newblock Video-Based Person Re-Identification via Self Paced Weighting.
\newblock In \emph{AAAI}, 2273--2280.

\bibitem[{Jiang et~al.(2021{\natexlab{a}})Jiang, Wang, Yi, Chen, Wang, Wang,
  Jiang, and Lin}]{jiang2021rain}
Jiang, K.; Wang, Z.; Yi, P.; Chen, C.; Wang, Z.; Wang, X.; Jiang, J.; and Lin,
  C.-W. 2021{\natexlab{a}}.
\newblock Rain-free and residue hand-in-hand: A progressive coupled network for
  real-time image deraining.
\newblock \emph{IEEE Trans. Image Process.}, 30: 7404--7418.

\bibitem[{Jiang et~al.(2019)Jiang, Wang, Yi, Wang, Gu, and
  Jiang}]{jiang2019atmfn}
Jiang, K.; Wang, Z.; Yi, P.; Wang, G.; Gu, K.; and Jiang, J. 2019.
\newblock ATMFN: Adaptive-threshold-based multi-model fusion network for
  compressed face hallucination.
\newblock \emph{IEE Trans. Multim.}, 22(10): 2734--2747.

\bibitem[{Jiang et~al.(2021{\natexlab{b}})Jiang, Gong, Liu, Cheng, Fang, Shen,
  Yang, Zhou, and Wang}]{jiang2021enlightengan}
Jiang, Y.; Gong, X.; Liu, D.; Cheng, Y.; Fang, C.; Shen, X.; Yang, J.; Zhou,
  P.; and Wang, Z. 2021{\natexlab{b}}.
\newblock Enlightengan: Deep light enhancement without paired supervision.
\newblock \emph{IEEE Trans. Image Process.}, 30: 2340--2349.

\bibitem[{Kaur, Kaur, and Kaur(2011)}]{kaur2011survey}
Kaur, M.; Kaur, J.; and Kaur, J. 2011.
\newblock Survey of contrast enhancement techniques based on histogram
  equalization.
\newblock \emph{International Journal of Advanced Computer Science and
  Applications}, 2(7).

\bibitem[{Lai et~al.(2017)Lai, Huang, Ahuja, and Yang}]{lai2017deep}
Lai, W.-S.; Huang, J.-B.; Ahuja, N.; and Yang, M.-H. 2017.
\newblock Deep laplacian pyramid networks for fast and accurate
  super-resolution.
\newblock In \emph{CVPR}, 624--632.

\bibitem[{Land(1977)}]{land1977retinex}
Land, E.~H. 1977.
\newblock The retinex theory of color vision.
\newblock \emph{Scientific american}, 237(6): 108--129.

\bibitem[{Ledig et~al.(2017)Ledig, Theis, Huszár, Caballero, Cunningham,
  Acosta, Aitken, Tejani, Totz, Wang, and Shi}]{8099502}
Ledig, C.; Theis, L.; Huszár, F.; Caballero, J.; Cunningham, A.; Acosta, A.;
  Aitken, A.; Tejani, A.; Totz, J.; Wang, Z.; and Shi, W. 2017.
\newblock Photo-Realistic Single Image Super-Resolution Using a Generative
  Adversarial Network.
\newblock In \emph{CVPR}, 105--114.

\bibitem[{Lee, Lee, and Kim(2012)}]{lee2012contrast}
Lee, C.; Lee, C.; and Kim, C.-S. 2012.
\newblock Contrast enhancement based on layered difference representation.
\newblock In \emph{ICIP}, 965--968.

\bibitem[{Li et~al.(2018)Li, Guo, Porikli, and Pang}]{li2018lightennet}
Li, C.; Guo, J.; Porikli, F.; and Pang, Y. 2018.
\newblock LightenNet: a convolutional neural network for weakly illuminated
  image enhancement.
\newblock \emph{Pattern Recognition Letters}, 104: 15--22.

\bibitem[{Liu et~al.(2014)Liu, Liu, Huang, and Bovik}]{liu2014no}
Liu, L.; Liu, B.; Huang, H.; and Bovik, A.~C. 2014.
\newblock No-reference image quality assessment based on spatial and spectral
  entropies.
\newblock \emph{SPIC}, 29(8): 856--863.

\bibitem[{Loh and Chan(2019)}]{Exdark}
Loh, Y.~P.; and Chan, C.~S. 2019.
\newblock Getting to know low-light images with the Exclusively Dark dataset.
\newblock \emph{Comput. Vis. Image Underst.}, 178: 30--42.

\bibitem[{Lv et~al.(2018)Lv, Lu, Wu, and Lim}]{Lv2018MBLLEN}
Lv, F.; Lu, F.; Wu, J.; and Lim, C. 2018.
\newblock MBLLEN: Low-Light Image/Video Enhancement Using CNNs.
\newblock In \emph{BMVC}, 220.

\bibitem[{Ma, Zeng, and Wang(2015)}]{ma2015perceptual}
Ma, K.; Zeng, K.; and Wang, Z. 2015.
\newblock Perceptual quality assessment for multi-exposure image fusion.
\newblock \emph{IEEE Trans. Image Process.}, 24(11): 3345--3356.

\bibitem[{Mittal, Soundararajan, and Bovik(2012)}]{mittal2012making}
Mittal, A.; Soundararajan, R.; and Bovik, A.~C. 2012.
\newblock Making a “completely blind” image quality analyzer.
\newblock \emph{IEEE Signal Process. Lett.}, 20(3): 209--212.

\bibitem[{{Papyan} and {Elad}(2016)}]{7327182}
{Papyan}, V.; and {Elad}, M. 2016.
\newblock Multi-Scale Patch-Based Image Restoration.
\newblock \emph{IEEE Trans. Image Process.}, 25(1): 249--261.

\bibitem[{Wang et~al.(2019)Wang, Zhang, Fu, Shen, Zheng, and
  Jia}]{wang2019underexposed}
Wang, R.; Zhang, Q.; Fu, C.-W.; Shen, X.; Zheng, W.-S.; and Jia, J. 2019.
\newblock Underexposed photo enhancement using deep illumination estimation.
\newblock In \emph{CVPR}, 6849--6857.

\bibitem[{Wang et~al.(2020)Wang, Chen, Wang, Liu, Satoh, Liang, and
  Lin}]{NightSurveillance}
Wang, X.; Chen, J.; Wang, Z.; Liu, W.; Satoh, S.; Liang, C.; and Lin, C. 2020.
\newblock When Pedestrian Detection Meets Nighttime Surveillance: {A} New
  Benchmark.
\newblock In \emph{IJCAI}, 509--515.

\bibitem[{Wang et~al.(2004)Wang, Bovik, Sheikh, and Simoncelli}]{wang2004ssim}
Wang, Z.; Bovik, A.~C.; Sheikh, H.~R.; and Simoncelli, E.~P. 2004.
\newblock Image quality assessment: from error visibility to structural
  similarity.
\newblock \emph{IEEE Trans. Image Process.}, 13(4): 600--612.

\bibitem[{Wei et~al.(2018)Wei, Wang, Yang, and Liu}]{wei2018deep}
Wei, C.; Wang, W.; Yang, W.; and Liu, J. 2018.
\newblock Deep Retinex Decomposition for Low-Light Enhancement.
\newblock In \emph{BMVC}, 155.

\bibitem[{Wu and Saito(2017)}]{wu2017interactive}
Wu, J.-H.; and Saito, S. 2017.
\newblock Interactive relighting in single low-dynamic range images.
\newblock \emph{ACM TOG}, 36(2): 1--18.

\bibitem[{Xu et~al.(2021{\natexlab{a}})Xu, Liu, Zhang, Guan, and
  Hu}]{xu2021rethinking}
Xu, X.; Liu, L.; Zhang, X.; Guan, W.; and Hu, R. 2021{\natexlab{a}}.
\newblock Rethinking data collection for person re-identification: active
  redundancy reduction.
\newblock \emph{Pattern Recognition}, 113: 107827.

\bibitem[{Xu et~al.(2021{\natexlab{b}})Xu, Wang, Wang, Zhang, and
  Hu}]{xu2021exploring}
Xu, X.; Wang, S.; Wang, Z.; Zhang, X.; and Hu, R. 2021{\natexlab{b}}.
\newblock Exploring Image Enhancement for Salient Object Detection in Low Light
  Images.
\newblock \emph{ACM Trans. Multim. Comput. Commun. Appl.}, 17(1s): 1--19.

\bibitem[{Yang et~al.(2020{\natexlab{a}})Yang, Zhu, Chen, Yan, Zhang, and
  Willis}]{yang2020mutualnet}
Yang, T.; Zhu, S.; Chen, C.; Yan, S.; Zhang, M.; and Willis, A.
  2020{\natexlab{a}}.
\newblock Mutualnet: Adaptive convnet via mutual learning from network width
  and resolution.
\newblock In \emph{ECCV}, 299--315. Springer.

\bibitem[{Yang et~al.(2020{\natexlab{b}})Yang, Wang, Fang, Wang, and
  Liu}]{yang2020fidelity}
Yang, W.; Wang, S.; Fang, Y.; Wang, Y.; and Liu, J. 2020{\natexlab{b}}.
\newblock From fidelity to perceptual quality: A semi-supervised approach for
  low-light image enhancement.
\newblock In \emph{CVPR}, 3063--3072.

\bibitem[{Yu, Yang, and Chen(2021)}]{yu2021towards}
Yu, W.; Yang, T.; and Chen, C. 2021.
\newblock Towards Resolving the Challenge of Long-tail Distribution in UAV
  Images for Object Detection.
\newblock In \emph{WACV}, 3258--3267.

\bibitem[{Yue et~al.(2017)Yue, Yang, Sun, Wu, and Hou}]{yue2017contrast}
Yue, H.; Yang, J.; Sun, X.; Wu, F.; and Hou, C. 2017.
\newblock Contrast enhancement based on intrinsic image decomposition.
\newblock \emph{IEEE Trans. Image Process.}, 26(8): 3981--3994.

\bibitem[{Zamir et~al.(2020)Zamir, Arora, Khan, Hayat, Khan, Yang, and
  Shao}]{zamir2020learning}
Zamir, S.~W.; Arora, A.; Khan, S.; Hayat, M.; Khan, F.~S.; Yang, M.~H.; and
  Shao, L. 2020.
\newblock Learning Enriched Features for Real Image Restoration and
  Enhancement.
\newblock In \emph{ECCV}, 492--511.

\bibitem[{Zhang et~al.(2011)Zhang, Zhang, Mou, and Zhang}]{zhang2011fsim}
Zhang, L.; Zhang, L.; Mou, X.; and Zhang, D. 2011.
\newblock FSIM: A feature similarity index for image quality assessment.
\newblock \emph{IEEE Trans. Image Process.}, 20(8): 2378--2386.

\bibitem[{Zhang et~al.(2020)Zhang, Di, Zhang, and Wang}]{zhang2020self}
Zhang, Y.; Di, X.; Zhang, B.; and Wang, C. 2020.
\newblock Self-supervised image enhancement network: Training with low light
  images only.
\newblock \emph{arXiv e-prints}, 2002.11300.

\bibitem[{Zhang et~al.(2021)Zhang, Guo, Ma, Liu, and Zhang}]{zhang2021beyond}
Zhang, Y.; Guo, X.; Ma, J.; Liu, W.; and Zhang, J. 2021.
\newblock Beyond Brightening Low-light Images.
\newblock \emph{International Journal of Computer Vision}, 129(4): 1013--1037.

\bibitem[{Zhang et~al.(2018)Zhang, Li, Li, Wang, Zhong, and
  Fu}]{zhang2018image}
Zhang, Y.; Li, K.; Li, K.; Wang, L.; Zhong, B.; and Fu, Y. 2018.
\newblock Image super-resolution using very deep residual channel attention
  networks.
\newblock In \emph{ECCV}, 286--301.

\bibitem[{Zhang, Zhang, and Guo(2019)}]{Zhang2019Kindling}
Zhang, Y.; Zhang, J.; and Guo, X. 2019.
\newblock Kindling the darkness: A practical low-light image enhancer.
\newblock In \emph{ACM MM}, 1632--1640.

\bibitem[{Zhong et~al.(2021)Zhong, Lu, Huang, Ye, Jia, and
  Lin}]{zhong2021grayscale}
Zhong, X.; Lu, T.; Huang, W.; Ye, M.; Jia, X.; and Lin, C.-W. 2021.
\newblock Grayscale enhancement colorization network for visible-infrared
  person re-identification.
\newblock \emph{IEEE Trans. Circuits Syst. Video Technol.}

\bibitem[{Zhu et~al.(2020)Zhu, Pan, Chen, and Yang}]{zhu2020eemefn}
Zhu, M.; Pan, P.; Chen, W.; and Yang, Y. 2020.
\newblock Eemefn: Low-light image enhancement via edge-enhanced multi-exposure
  fusion network.
\newblock In \emph{AAAI}, volume~34, 13106--13113.

\end{thebibliography}
\end{document}